\documentclass[review]{elsarticle}

\usepackage{lineno,amssymb,graphicx}
\usepackage{amsmath}
\usepackage{bm}
\usepackage{epstopdf}%
\usepackage{bm}
\usepackage{multirow}
\usepackage{booktabs}
\modulolinenumbers[5]

\biboptions{numbers,sort&compress}
\usepackage[utf8x]{inputenc}

\usepackage{nameref,hyperref}

\usepackage{array}

\newcolumntype{+}{!{\vrule width 2pt}}

\newlength\savedwidth

\usepackage[aboveskip=1pt,labelfont=bf,labelsep=period,justification=raggedright,singlelinecheck=off]{caption}

\makeatletter
\def\ps@pprintTitle{%
 \let\@oddhead\@empty
 \let\@evenhead\@empty
 \def\@oddfoot{\centerline{Preprint submitted to PLoS One. June 1, 2020}}%
 \let\@evenfoot\@oddfoot}
\makeatother

\begin{document}

\begin{frontmatter}

\title{An embedded system for the automated generation of labeled plant images to enable machine learning applications in agriculture}

\author{M.A. Beck$^{a,b,*}$, C.-Y. Liu$^{c}$, C.P. Bidinosti$^{a,b,c}$, C.J. Henry$^{b}$, C.M. Godee$^{d}$,\break M. Ajmani$^{a,b}$}
\address{$^a$Departmnet of Physics, Univeristy of Winnipeg, Winnipeg, MB, R38 2E9, Canada}
\address{$^b$Departmnet of Applied Computer Science, Univeristy of Winnipeg, Winnipeg, MB, R38 2E9, Canada}
\address{$^c$Departmnet of Electrical and Computer Engineering, Univeristy of Manitoba, Winnipeg, MB, R3T 5V6, Canada}
\address{$^d$Departmnet of Biology, Univeristy of Winnipeg, Winnipeg, MB, R38 2E9, Canada}
\address {$^*$Corresponding authour, E-Mail address: m.beck@uwinnipeg.ca}

\begin{abstract}
A lack of sufficient training data, both in terms of variety and quantity, is often the bottleneck in the development of machine learning (ML) applications in any domain.  For agricultural applications, ML-based models designed to perform tasks such as autonomous plant classification will typically be coupled to just one or perhaps a few plant species. As a consequence, each crop-specific task is very likely to require its own specialized training data, and the question of how to serve this need for data now often overshadows the more routine exercise of actually training such models. To tackle this problem, we have developed an embedded robotic system to automatically generate and label large datasets of plant images for ML applications in agriculture.  The system can image plants from virtually any angle, thereby ensuring a wide variety of data; and with an imaging rate of up to one image per second, it can produce lableled datasets on the scale of thousands to tens of thousands of images per day. As such, this system offers an important alternative to time- and cost-intensive methods of manual generation and labeling.  Furthermore, the use of a uniform background made of blue keying fabric enables additional image processing techniques such as background replacement and plant segmentation. It also helps in the training process, essentially forcing the model to focus on the plant features and eliminating random correlations. To demonstrate the capabilities of our system, we generated a dataset of over 34,000 labeled images, with which we trained an ML-model to distinguish grasses from non-grasses in test data from a variety of sources. We now plan to generate much larger datasets of Canadian crop plants and weeds that will be made publicly available in the hope of further enabling ML applications in the agriculture sector.
\end{abstract}


\end{frontmatter}


\section*{Official Publication}
This work was officially published under \url{https://journals.plos.org/plosone/article?id=10.1371/journal.pone.0243923}. For citing this work please use the citation provided by PLOS One. 

\section{Introduction}

A review of the recent literature shows there is great optimism that advances in 
sensors \cite{Vazquez2016,Narvaez2017,Antonacci2018,KHANNA2019218}, 
robotics \cite{Oberti2016,Bechar2016,Bechar2017,Duckett2018, 20193426318, RELFECKSTEIN2019100307}, 
and machine learning \cite{Lobet2017,Waldchen2018,Liakos2018,patricio2018computer,kamilaris2018deep,JHA20191} 
will bring new innovations destined to increase agricultural production and global food security.
Whether one speaks more broadly of  precision agriculture, digital agriculture or Agriculture~4.0 (in reference to the anticipated fourth agricultural revolution), the confluence of these technologies in particular could lead, for example, to automated methods of weeding,  disease evaluation, plant care, and phenotyping \cite{JHA20191,Binch2017,Bah2018,Bosilj2018, barbedo2013digital, Fahgen2015,Singh2016,Shakoor2017,Gehan2017,Giuffrida2018,Tardieu2017}.
Such capabilities would increase crop yields and expedite breeding programs, while minimizing inputs (e.g.\ water, fertilizer, herbicide, pesticide) and reducing the impact on the environment.

Prototypes of autonomous vehicles performing farming tasks in the field exist
already \cite{20193426318,RELFECKSTEIN2019100307,7487720,doi:10.5772/62059}.
However, putting the ``brains'' into such agents is still a hard challenge
and success is limited to a crop's specifics and the task at hand.  
Machine learning (ML) utilizing convolutional neural networks (CNNs)  holds great promise for image-based location and identification tasks in agriculture. 
The capabilities of CNNs  have improved vastly in recent years \cite{ILSVRC15,LeCun2015,Kaiming2015} and
are now used as solutions to previously difficult problems such as object detection within images \cite{NIPS2015_5638}, facial recognition \cite{Taigman_2014_CVPR}, automatic image annotation \cite{Vinyals_2015_CVPR}, self-driving cars \cite{BojarskiTDFFGJM16} and automated map production \cite{Henry2019}.

While there are many different CNN architectures and training methods, a general rule of thumb is the following: A
model's capability to identify objects in previously unseen data (called generalizing) depends significantly on the amount of data the model \emph{has seen} during training \cite{ILSVRC15, unreasonable2017}.  As a result, an inadequate amount of  high-quality training data---in particular, labeled data---is often the bottleneck in developing ML-based applications, a fact underscored by many authors working in plant sciences and agriculture \cite{Lobet2017,Waldchen2018,Liakos2018,Binch2017,Bah2018,Bosilj2018,Fahgen2015,Singh2016,Shakoor2017,Gehan2017,Giuffrida2018,Tardieu2017}.
This problem is magnified by the circumstance that each application is likely to require its own specific training data, especially given the very wide variety of plant appearances, e.g. tillering versus ripening, healthy versus  diseased, crop versus weed.
For example, training CNNs to distinguish oats from their wild counterpart---which are responsible for an annual loss of up to \$500 million in the Province of Manitoba
alone\footnote{According to \href{https://www.gov.mb.ca/agriculture/crops/weeds/wild-oats.html}{https://www.gov.mb.ca/agriculture/crops/weeds/wild-oats.html}} ---would certainly require a qualitatively and quantitatively rich dataset of labeled images of all variants. 

The need for labeled training data is often satisfied by manual annotation, which is typically achieved through one of two ways.  If the classification problem is common knowledge, it can be crowdsourced, as is done through  platforms, such as Mechanical Turk \cite{buhrmester2016amazon} and ReCaptcha \cite{schenk2009crowdsourcing}.
Conversely, if the classification problem requires expert knowledge, crowdsourcing will not be reliable and annotation must be performed by experts only.  Both methods have been suggested for labeling plant images \cite{Waldchen2018,Singh2016,Gehan2017,Giuffrida2018}, and 
although there are tools available to ease the process \cite{russell2008labelme,rapson2018reducing,dutta2019via},
manual annotation is both cost- and time-intensive and usually leads to comparably small datasets in the magnitude of a couple of thousands images. As a workaround to having large, labeled datasets, several strategies, such as transfer learning with smaller labeled data sets \cite{Waldchen2018,Ubbens2017,Ubbens2018}  or unsupervised learning with unlabeled data \cite{Bah2018}, are being explored.  Given the preference for large, labeled datasets, data augmentation is also being employed and ranges from simple modifications of existing images (e.g. rotation, translation, flipping, and scaling) \cite{Waldchen2018} to generating synthetic images \cite{Lobet2017,Waldchen2018,Ubbens2018}.  

In an effort to produce large quantities of high-quality training data for machine learning applications in agriculture, we have developed an embedded system to automatically generate and label images of real plants.  This system---henceforth referred to as EAGL-I (\textbf{E}mbedded \textbf{A}utomated \textbf{G}enerator
of \textbf{L}abeled \textbf{I}mages)---is, in a nutshell, a robotically moved camera that takes pictures of known plants at known locations from a large variety of  known positions and angles. Since we have full information and control over where on the image the plants are located, we can automatically identify and label them.
As a result, EAGL-I can generate labeled data at the rate of thousands to tens of thousands of images per day, with minimal human interaction and no dependence on crowdsourcing or expert knowledge.

\begin{figure}[tb]
\begin{center}\includegraphics[width=1.0\columnwidth]{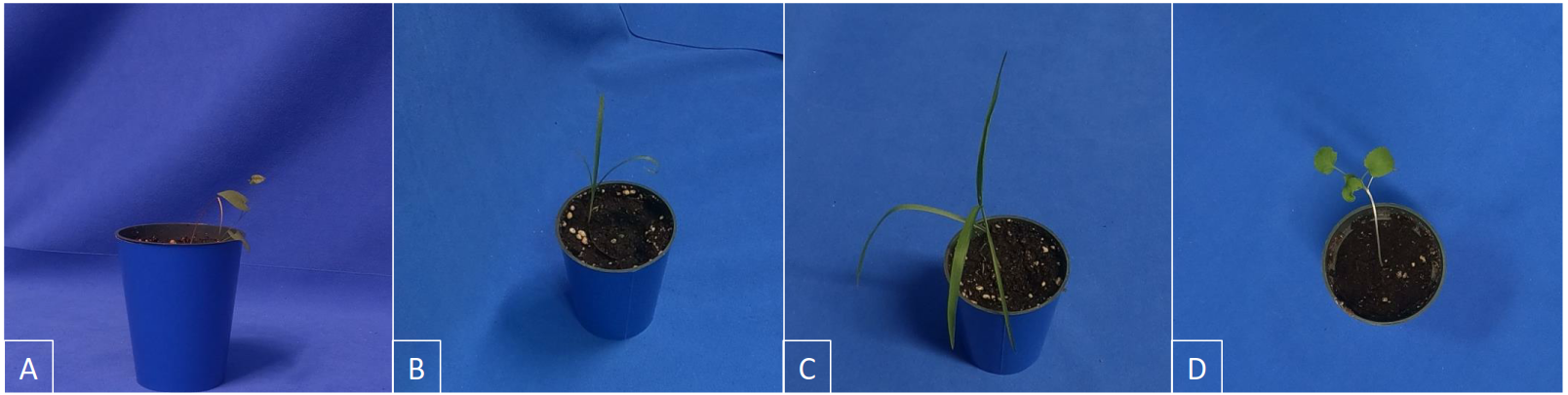}\end{center}

\caption{{\bf Example images taken by EAGL-I.}
A: Wild buckwheat in a profile shot. B-C: Yellow foxtail and barnyard grass in oblique angles. D: Canola in an overhead shot. Blue keying fabric is used as background.}
\label{fig1}
\end{figure}

While there are many examples in the literature of plant imaging 
systems already \cite{crimmins2008monitoring, doi:10.1111/tpj.12131,  doi:10.1111/j.1469-8137.2005.01609.x, jansen2009simultaneous, CHENE2012122, dobrescu2017yourself, Minervini2015PRL, doi:10.1111/tpj.13472,  BAI2016181,  JIANG201657,  BARKER201674, 
jimenez2018high, 10.1007/s00138-015-0670-5, 10.1371/journal.pone.0196615},
their primary purpose has been to capture and compare phenotypic information and growth metrics. This is typically achieved through overhead shots only and requires close-to-zero variance in imaging conditions  to ensure a high accuracy in extracting plant characteristics.  This is at odds with the type of datasets needed to train machine learning algorithms for plant classification.  In this case, one is interested in a \emph{rich} dataset, with a wide variety of images falling under the same label. Variety can be achieved through differences in used parameters, such as imaging angle (Fig \ref{fig1}), camera-to-plant distance, lighting conditions, time of day, growth stage, and the use of different plants of the same cultivar or species.  One must also include different plants with different growing characteristics.
For example corn (a fast-growing, tall grass) is very different, say, compared to dandelion (a ground-hugging rosette), but one still needs examples of both (and indeed others) in the same training set to identify crop versus weed with the highest possible accuracy.
EAGL-I has the capabilities to incorporate all these differences and is, to the best of our knowledge, the only imaging system fully dedicated to the goal of generating machine learning datasets for plant classification.

The contributions of this paper are the following:
\begin{itemize}
\item We designed an imaging system to create labeled
datasets for training machine learning models
\item This system has a high imaging rate and autonomously labels the imaged
plants, offering an alternative to cost-intensive manual labeling.
\item The system can image plants from any angle and at different distances,
thus, producing the variety needed for training datasets
\item A wide variety of plants can be imaged and there is full freedom in
their arrangement in the coverable volume
\item As a proof of concept, we generated a dataset of different weeds commonly found in the Province of Manitoba, 
trained a CNN with it, and evaluated the resulting model on previously unseen data
\end{itemize}

The rest of the paper is structured as follows. Section \ref{sec:System-Overview}
describes the EAGL-I's parts, specifications, and mode of
operation. Section \ref{sec:Data-Production} describes data generation and defines the imaging rate of EAGL-I.
It also lists the parameters we used in production to generate a training
dataset. In Section \ref{sec:The-Weed-Dataset}, we characterize that dataset
and use it to train a CNN to distinguish dicots from monocots. 
Section \ref{sec:Conclusion-and-Future}
concludes the paper and discusses planned improvements to the system and future work.

\begin{table}[!ht]
\caption{{\bf System overview}}
\begin{center}
\begin{tabular}{|>{\centering}p{2cm}|>{\centering}p{2cm}|c|>{\centering}p{4cm}|}
\hline 
Part & Brand & Model & Specs \tabularnewline
\hline 
\hline 
X, Y, and Z Actuators & Macron Dynamics & MSA-628 & Volume covered (mm$^3$): \(1150\times 840\times 718\) \tabularnewline
\hline 
Planetary Gearbox & Servo-Elements & MPS-60-005 & 5:1 Ratio \tabularnewline
\hline 
Stepper Motors (XYZ) & Servo-Elements & ST24-1.8-297 & NEMA 24, 24~V DC 2.8~A\break 1.8° step angle\break 2.7 N$\,$m rated torque.\break Stepper drives integrated.\tabularnewline
\hline 
Controller of X, Y, Z Actuators & Arduino & Uno Rev3 & Microcontroller: ATmega328P\break Clock Speed: 16 MHz\break Max. Pulse Rate:\break 4000 pulses/second\tabularnewline
\hline 
RGB-Camera & GoPro & Hero 7 Black & Res: $4000 \times 3000$ px\break Used in linear mode, no zoom: FOV = 98.7°\break File format: jpg\tabularnewline
\hline 
Servo Motors (pan-tilt) & Dynamixel & MX-28T & 11.1-14.8~V, 1.4~A\break 0.088° step angle\break 2.5 N$\,$m stall torque \tabularnewline
\hline 
AC/DC Converter & Mean Well USA & LRS-350-36 & Output: 36~V, 9.7~A, 350~W \tabularnewline
\hline 
\end{tabular}
\end{center}
\label{tab:System-Overview}
\end{table}

\section{System Overview\label{sec:System-Overview}}

Table \ref{tab:System-Overview} gives an overview on the EAGL-I hardware.

\begin{figure}[!h]
\begin{center}\includegraphics[width=1.0\columnwidth]{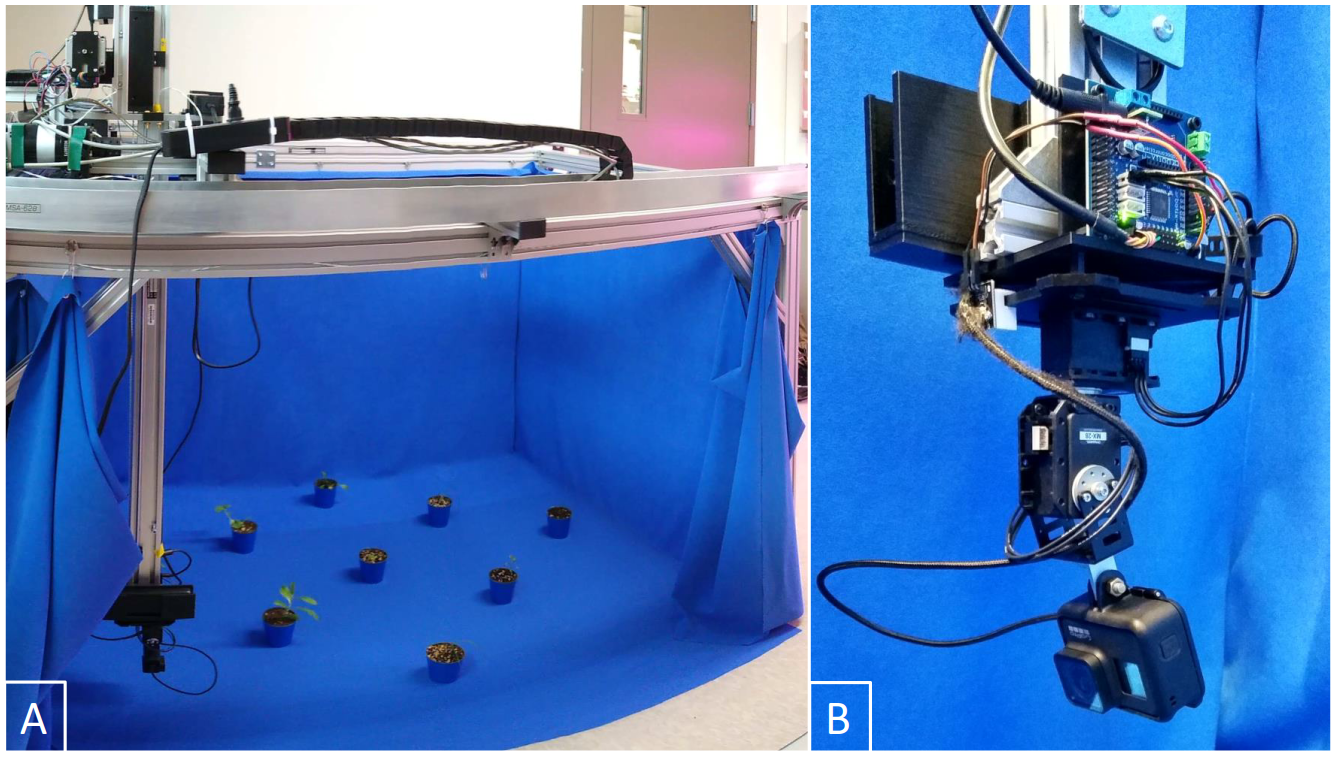}\end{center}
\caption{{\bf The EAGL-I system.} 
A: Full view with blue keying fabric pulled back to show the imaging volume. B: Close-up view of the gantry head carrying the pan-tilt system, the camera, and a powerbank.}
\label{fig2}
\end{figure}

The system is setup in a gantry configuration (Fig \ref{fig2}), such that the gantry head can be moved 
in all three dimensions of a volume measuring\break 115 x 84 x 71 cm$^3$.
Two actuators per axis provide movement in the X-Y-plane and a fifth actuator raises or lowers the gantry head.
For safety and repeatability, we equipped the actuators with limit and homing switches. 
The normally closed limit switches prevent the actuators to move beyond their bounds. When the switches
trigger (or lose power) the whole systems shuts off immediately and
until a manual reset. The homing switches counteract possible drifts or slips of the actuators. 
An Arduino Uno controls the gantry system's actuators, with
power supplied by a 350-W AC/DC converter. 

On the gantry head we attached a pan-tilt system followed by an RGB camera. 
An Arduino-compatible micro-controller powers and
controls the pan-tilt system via two servo motors, allowing the camera to be rotated through any combination of azimuthal and polar angles (360$^\circ$ pan, 180$^\circ$ tilt). The camera itself is powered by a commercial 20-A$\,$h power bank that can support its imaging process for over 8~hours and is easily swapped out.

\section{Data Production\label{sec:Data-Production}}

The two main contributions here are the duration of the robotic movement and the image processing 
time of the camera, each of which are discussed separately below.

\subsection{Robotic Movement}

The camera is moved by the xyz-gantry and the pan-tilt-subsystem.
Since panning and tilting the camera happens in parallel to the movement
in x, y, and z (and is almost always faster), we can neglect that 
contribution for the imaging rate. We control the actuators
close to the maximal pulse-rate the Arduino Uno can output (4000 pulses per second). 
This translates into a movement speed of
\begin{equation}
v=p_{r}\cdot d\cdot s\cdot r\cdot m=p_{r}\cdot0.105\cdot m,
\end{equation}
where $p_{r}$ is the pulse rate, $d=105$ is the distance traveled per revolution
of the actuator in millimeters, $s=1.8/360=0.005$ is the fraction
of a full revolution made by 1 step of the stepper motors, $r=0.2$
is the gearbox's reduction ratio, and $m$ is a factor determined
by the stepping mode. For full-stepping mode $m=1$, whereas half-stepping
means $m=0.5$. The controller uses a linear acceleration and deceleration
profile to ease in and out of the actuators' movements. Overall, then, 
we have a nearly linear proportional relationship between pulse rate
and travel speed. Furthermore, all three axis can be moved in parallel or one after each other. 

When the camera is moved to a new position and orientation, 
it is useful to pause before proceeding to trigger it to take an image. 
This allows vibrations to settle down and not doing so might result in blurry images, especially
when using longer exposure times.

\begin{figure}[!h]
\begin{center}
\begin{center}\includegraphics[width=0.8\columnwidth]{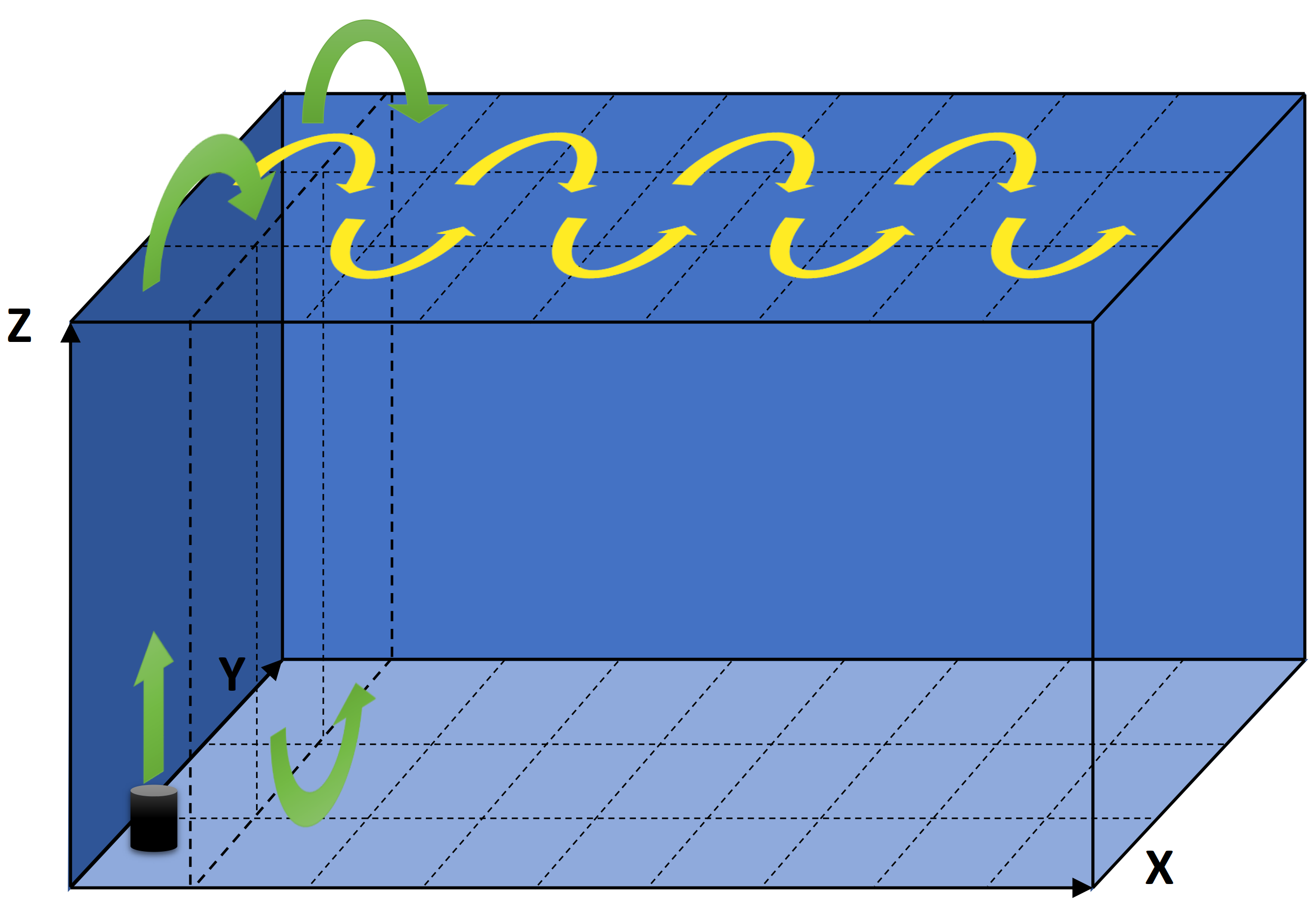}\end{center}
\end{center}
\caption{{\bf Path of gantry head.}
Movement of the gantry head in a zig-zag motion through columns and
slabs of the coverable volume, starting in the bottom left near corner.
The yellow arrows depict the motion from one slab to the next, nested
inside those movements are the motions from one column to the next,
depicted by the green arrows (only shown for the first slab).}
\label{fig3}
\end{figure}

When going through many different camera positions in sequence, the
order in which those positions are visited is of equal, if not even
higher, importance than the speed with which the camera is moved. To
obtain a general optimal solution one would have to solve a three-dimensional
traveling salesman problem (TSP), which is a well-known NP-hard problem
in combinatorial optimization. In our typical application, we would
have to solve the TSP for thousands of different positions.
While still feasible, we settled for a nested zig-zag algorithm,
as depicted in Fig \ref{fig3}, which offers a straightforward method to keep travel times between 
successive camera positions short.

The cuboid-shaped volume through which the gantry system can move the camera
is divided into slabs of equal width along its X-axis. Those slabs
are all subdivided into equally wide columns along the Y-axis.
Now, starting at the bottom of the first column (containing the
coordinate system's origin), we move the gantry head to the
position inside that column with the smallest Z-value. From there
we move upwards through the positions with the next-largest Z-values inside
that column (ties in Z-values are resolved arbitrarily). Note that
small movements in X- and Y-direction are still happening, but are
limited by the columns boundaries. We keep moving upwards until reaching the 
highest position inside the column. From there
we continue to the next column in positive Y-direction and reverse
the procedure: we start with the position having the largest Z-value
and descend through the column. We keep zig-zagging through the first slab's columns
until we reach the end of its last column. From
there we move to the second slab in positive X-direction. We continue
a zig-zag motion working our way through the columns, but this time,
when we change columns we move in negative Y-direction, until having
traversed the entire second slab. We continue those zig-zag motions
from slab to slab, until each position was visited.

\subsection{Imaging Process}

The imaging process is initiated by sending an HTTP request to the
camera over WiFi. The delay to send and process the request is
negligible (of order of a few milliseconds) and thus is of no concern
for the imaging rate. The time to perform the imaging itself depends
on the camera settings and lighting conditions. In our indoor setup,
without additional light sources and a maximal ISO of 200, the camera
needs approximately 2.7 seconds to take an image. Allowing a higher exposure index 
would reduce that time, but also introduce grain to the image.
Additional lighting will reduce the exposure time, but is presently not a main concern.

Images can be downloaded from the camera via a USB or WiFi connection.
In either case, one can retrieve each image directly after it has been captured or retrieve all images
in bulk after the system went through each of its positions. Retrieving
the images in bulk decouples the imaging procedure from retrieving
the data. By doing so, any delays or problems when transferring the images
does not interfere with collecting the images. For the
sake of automation, we value image collection higher than the data
retrieval, since data generation takes much longer than its retrieval
and thus is harder to repeat.

\begin{figure}

\begin{center}\includegraphics[width=1.0\columnwidth]{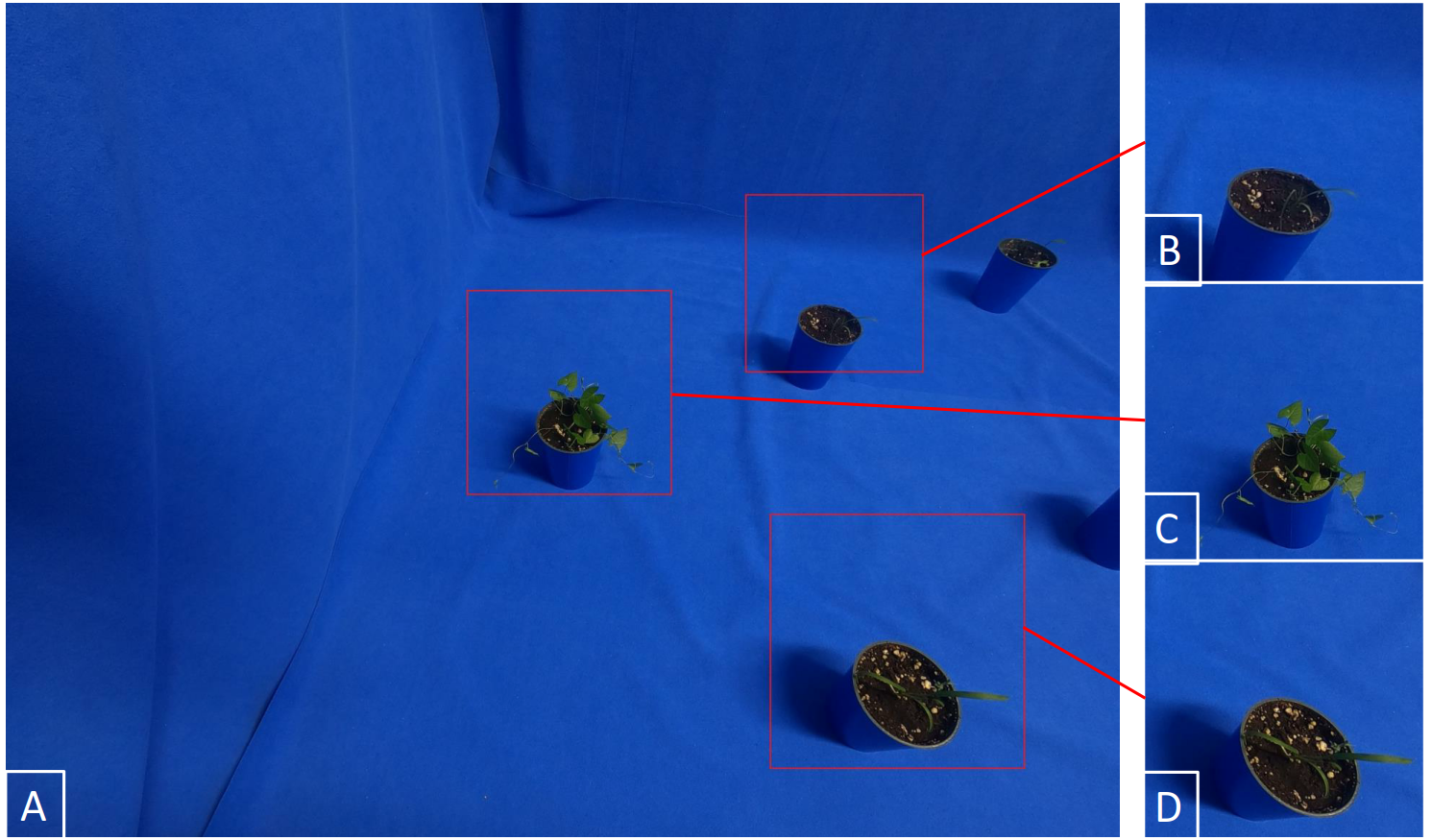}\end{center}

\caption{{\bf Master image and cropped images.}
A: Original master image taken by EAGL-I. B-- D: three subimages cropped out from it. 
Note that the cropped images have different dimensions, whereas we present them here at the same size.}
\label{fig4}
\end{figure}

Depending on the application, an easy way to increase imaging rate
is by cropping several subimages from a single image taken a given position. In our application
(generating single plant training data) this is a valid approach and can increase
imaging rate up to one order of magnitude. Cropping out subimages results in
different image sizes, which could be considered
a drawback for some applications, but is rarely so in machine learning.
Fig \ref{fig4} shows an example of cropping several
images from a master image.

\subsection{Production Settings}

We define average production times $t_m$ and $t_s$ for master- and subimages, respectively, as follows: 

\begin{equation}
t_{m}=\frac{t_{p}+t_{d}}{N_m}
\label{tm}
\end{equation}
\begin{equation}
t_{s}=\frac{t_{p}+t_{d}+t_{c}}{N_s} \, ,
\label{ts}
\end{equation}
where $t_{p}$ is the total time required to produce $N_m$ master images (including
robotic movements), $t_{d}$ is the time to bulk download all master images from the camera to the computer, and  $t_{c}$ is the time required to crop out  a total of $N_s$ subimages from the master images.   

To create a training dataset, we have performed runs with the system on a daily basis under the settings listed
in Table \ref{tab:Production-Settings}. This resulted in $t_m \sim 7$~s and $t_s \sim 4.8$~s. 
Those settings are conservative and we have achieved during
testing $t_s < 1$~s.
Imaging at such fast rates comes at a cost of image quality, however.
First, the shorter exposure time increases the ISO needed, which in turn
introduces grain to the image. Second, to achieve maximal imaging
rates, we have to pack plants in a tighter arrangement under the system.
That can lead to overlap in the bounding boxes, i.e. meaning there
are cases in which we can see plant material of neighboring plants
in the images. Both points have to be accounted for, when using
the data as training sets in machine learning. Higher grain in the image
masks detailed features, 
and plant material from neighboring plants bring in unwanted features
that do not correlate with the actual plant in the image. Image quality
and imaging speed are two defining factors for the datasets that can
be produced by EAGL-I and often have to be traded off for one another.

\begin{table}[!htb]
\begin{center}
\caption{{\bf Production Settings}}
\begin{tabular}{|c|c|}
\hline 
Setting & Value\tabularnewline
\hline 
\hline 
Locations Imaged & 9\tabularnewline
\hline 
Parallel X, Y, Z Movement & No\tabularnewline
\hline 
Peak Pulse Rate & 3000 pulses$\cdot\text{s}^{-1}$ \tabularnewline
\hline 
Acceleration Rate & 10000 pulses$\cdot\text{s}^{-2}$\tabularnewline
\hline 
Stepping Mode & Half-steps\tabularnewline
\hline 
Pause before Camera Trigger & 3 seconds\tabularnewline
\hline 
Routing Algorithm & Nested Zig-Zag\tabularnewline
\hline 
Maximal ISO & 200\tabularnewline
\hline 
Imaging Time & Approx. 2.7 seconds\tabularnewline
\hline 
Additional Lighting  & None\tabularnewline
\hline 
Image Download & WiFi, In bulk\tabularnewline
\hline 
\noalign{\vskip\doublerulesep}
\hline 
Total Images & 2149\tabularnewline
\hline 
Total Subimages & 3494\tabularnewline
\hline 
Time for Imaging $t_{p}$ & 3 hours, 25 minutes\tabularnewline
\hline 
Download Time $t_{d}$ & 46 minutes\tabularnewline
\hline 
Cropping Time $t_{c}$ & 34 minutes\tabularnewline
\hline 
Imaging Rate (Images) $I_{r}$ & Approx. 7 s/image\tabularnewline
\hline 
Imaging Rate (Subimages) $I_{c}$ & Approx. 4.8 s/image\tabularnewline
\hline 
Size on Disk & 8-9 GB\tabularnewline
\hline 
\end{tabular}
\label{tab:Production-Settings}
\end{center}
\end{table}

\subsection{Cropping and Labeling Subimages}

Different methods are available to us for cropping out a single plant from a master image.
In the following we give a roadmap for two approaches based on image processing and CNNs, respectively. 
We chose for our system a third approach, instead, that relies on spatial information alone. 

An image processing approach relies on color differences between
the plant, the soil, and the image's background. With segmentation
algorithms we could identify the plants inside the image
and construct a minimal bounding box around it. We describe a similar process
in Subsection \ref{subsec:postprocessing}. In a second step the
segments would have to be matched to the plants' known positions to assign the correct label. 

Alternatively one could consider machine learning techniques themselves for cropping
and labeling subimages. This approach, however, can only be applied
once a sufficiently trained model is available. Here a two-step procedure
could be employed: First, a model is trained to define bounding boxes
in the image for each plant. These bounding boxes would again be matched
to the plants' known positions for labeling. Now, a second
model could be bootstrapped, that not only finds bounding boxes, but
also labels them by recognizing the plants shown. Keeping in
mind, that creating such models is ultimately the purpose
of EAGL-I, we encounter a ``chicken or egg'' problem.

In the case that there are more than one plant captured in one image,
both approaches mentioned above have to rely on the plants' spatial information
at one point or another to correctly match labels with subimages.
Only after achieving the goal, which EAGL-I was built to solve, we
can discard spatial information completely, while still correctly
labeling subimages. On the other side, spatial information is always
available to us and is \emph{sufficient} for cropping and labeling subimages.
This motivates the purely geometric approach we have implemented into our
system. It calculates the plants' coordinates inside the image from their known relative position and angle to the camera.
As a result, labeling sub-images becomes trivial.  Furthermore, the method is robust, as we do not have
to rely on the stability of an image processing pipeline or a machine
learning algorithm's accuracy.

To calculate the bounding box around
the plant we define a sequence of linear transformations that match
the plant's real-world coordinates (world frame) with
the plant's xy-position inside the image (image frame).
The net transform is 
\begin{equation}
T=T_{w2c}\cdot T_{c2i} \, .
\label{transform}
\end{equation}
Here $T_{w2c}$ is the linear transformation from world frame to camera frame,
i.e. a frame in which the camera is the origin pointing in positive
x-direction. Thus, the linear transformation $T_{w2c}$ consists of a translation,
depending on the gantry head position and the displacement due to
the pan-tilt system, and a rotation due to panning and tilting the
camera. The transformation $T_{c2i}$ converts the camera frame to
the image frame, meaning that the objects inside the camera's field
of view are being projected on the xy-coordinates of the image. For
this we calculate bearing and elevation of the object's position from
the camera. Using these angles we map the object to xy-coordinates (given in pixels),
depending on the camera's resolution and field of view. To calculate the object's
size in the image frame we calculate its subtended angle from the
camera. To this end, we replace, for calculations, the plant by
a sphere with radius large enough that the plant is contained inside
of it. For full details on these transformations,
we refer to our code in \cite{EAGLIcode}.

Given that we place plants on the floor (meaning the z-coordinate
is known), we can also invert the projection $T_{c2i}$ and the transform
$T_{w2c}$ to map the position and size of objects in image frame back
to world frame. This inversion effectively allows us to determine
any (sufficiently flat) object's x- and y-position from a single overhead
image taken by the system itself.

We want to point out that following a geometric approach
to locate the plants comes with its own challenges: It relies on precise
and accurate movements of the camera and location of the targets.
Accuracy and precision of our robot's movements turned out to be sufficient
for this approach. To achieve good positioning of the targets, we
measured and marked 12 target locations that we use repeatedly. The
system can also generate new target locations and mark
them with a laser. This allows us to not be limited to a fixed set
of positions. A second challenge to a geometric approach are lens
distortions, i.e. deviations from a perfect rectilinear projection
from camera frame to image frame. Such distortions usually appear on the image frame's
margins. We countermeasure those
drawbacks by using relatively large spheres to approximate the plants
imaged. Other countermeasures would be to measure the distortions and use software correction  
before cropping the subimages, or to simply not use subimages
that lie too close to the image's margins, or to use digital zoom that
effectively reduces the field of view to an area with only negligible lens distortions.

\subsection{Image Postprocessing\label{subsec:postprocessing}}

\begin{figure}[!h]

\begin{center}\includegraphics[width=1.0\columnwidth]{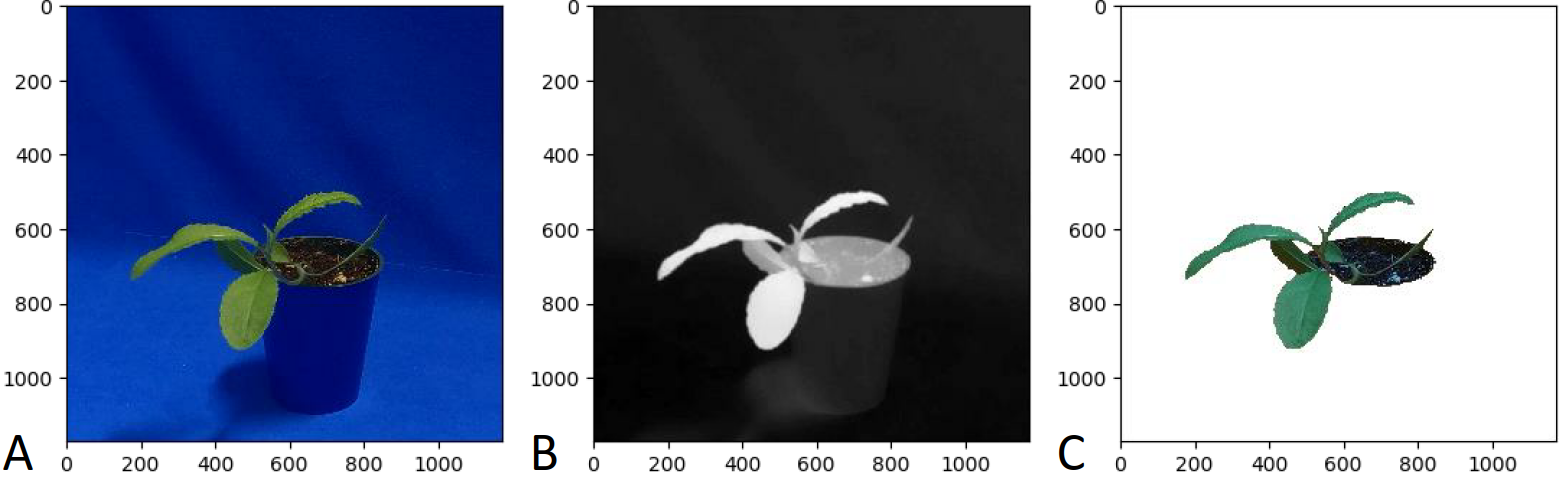}\end{center}

\caption{{\bf Background removal}
A: Original Image captured by EAGL-I. B: The originals blue-yellow channel as a grayscale image. C: Keyed out image. The background is removed by defining a threshold for the blue-yellow values. All pixels below that threshold are masked out.}
\label{fig5}
\end{figure}

As mentioned above, EAGL-I produces images against a neutral blue background. 
This enables
and simplifies image processing techniques. To demonstrate that, we
performed a simple color-based background removal on the example in
Fig \ref{fig5}. To this end, we used the PlantCV library for Python
\cite{gehan2017plantcv}, which itself is based on OpenCV \cite{bradski2008learning}.
In a first step we convert the image from the usual RGB color space
to the CIELAB color space, in which the \emph{b}-channel ranges from
low values for blue pixels to high values for yellow pixels. Fig
\ref{fig5}b shows the \emph{b}-channel of our example as
a grayscale image, the blue background appears dark, whereas the plant
and soil are bright shades of gray. With a binary threshold-filter
based on this channel we keyed out the plant as shown in Fig \ref{fig5}c.
Additional filters can be applied to remove artifacts and to smooth
the edges of the thresholding operation (e.g. dilating, filling
holes, eroding, Gaussian blur).

Since the camera positioning can be repeated precisely, a second technique
to key out the plant also becomes available: background subtraction. For
this technique a second picture is taken from the same position
and angle but without the plant. This image, that contains the background
only, can be subtracted from the image containing the plant,
leaving the plant itself.

Further image processing can be employed to remove the dark soil from
the green plant or to extract morphological information. Those techniques
are widely deployed in the area of (high throughput) phenotyping.
For those techniques we refer to \cite{gehan2017plantcv} and PlantCVs
online documentation.

\section{The Weedling Dataset\label{sec:The-Weed-Dataset}}

As proof of concept we have generated a labeled dataset of seedlings of eight weeds 
that are common in Manitoban fields. This dataset \cite{weedling2020} allows
us to test systems that lie downstream in the development pipeline, in
particular databases and the training of machine learning algorithms.

We chose weed species as targets, as they are of general interest and can be found amongst virtually every cash-crop in the field. 
The reasons to focus on a rather early growing stage are several. 
Using seedlings allows us to grow more individuals in rotation, discarding older plants for newer ones. This results in a 
richer dataset, compared to imaging a smaller number of individuals over their full life cycle. Furthermore, we can image more plants 
in parallel, thus achieving a higher imaging rate, if they are small. 
Lastly, a rather important and pragmatic argument is that the 
identification (and eradication) of weeds is most critical in the early stages of crop growth when plants are small and a canopy has not yet developed.

To generate the dataset we used the production settings as given in
Table \ref{tab:Production-Settings}. In 10 runs we generated 34,666
subimages of weeds in a total imaging time of 47 hours and 30 minutes.
Setting up the system to perform a single run requires personal attendance
of roughly 15 minutes, after which the system continues autonomously
and does not need further supervision. All images were taken in
front of the blue background (Figs.~\ref{fig1}, \ref{fig2} and~\ref{fig4}) 
to ease image processing and segmentation.
Furthermore, the uniform background helps in the training process to
focus the model on the plant features and eliminate random correlations.
Table \ref{tab:Weed-Dataset} gives an overview on the dataset's characteristics.

\begin{table}[!ht]
\caption{{\bf The Weedling Dataset}}
\begin{center}
\begin{tabular}{|c|c|}
\hline 
Weed & Number of Images*\tabularnewline
\hline 
\hline 
Echinochloa crus-galli (Barnyard Grass) & 8621\tabularnewline
\hline 
Cirsium arvense (Canada Thistle) & 4706\tabularnewline
\hline 
Brassica napus (Volunteer Canola) & 6723\tabularnewline
\hline 
Taraxacum officinale (Dandelion) & 4797\tabularnewline
\hline 
Persicaria spp. (Smartweed) & 870\tabularnewline
\hline 
Fallopia convolvulus (Wild Buckwheat) & 4621\tabularnewline
\hline 
Avena fatua (Wild Oat) & 1218\tabularnewline
\hline 
Setaria pumila (Yellow Foxtail) & 3110\tabularnewline
\hline 
\noalign{\vskip\doublerulesep}
\hline 
Total & 34,666\tabularnewline
\hline 
\end{tabular}
\begin{flushleft} *Variations are due to different germination success
\end{flushleft}
\end{center}
\label{tab:Weed-Dataset}
\end{table}

\begin{figure}[!h]

\begin{center}\includegraphics[width=1.0\columnwidth]{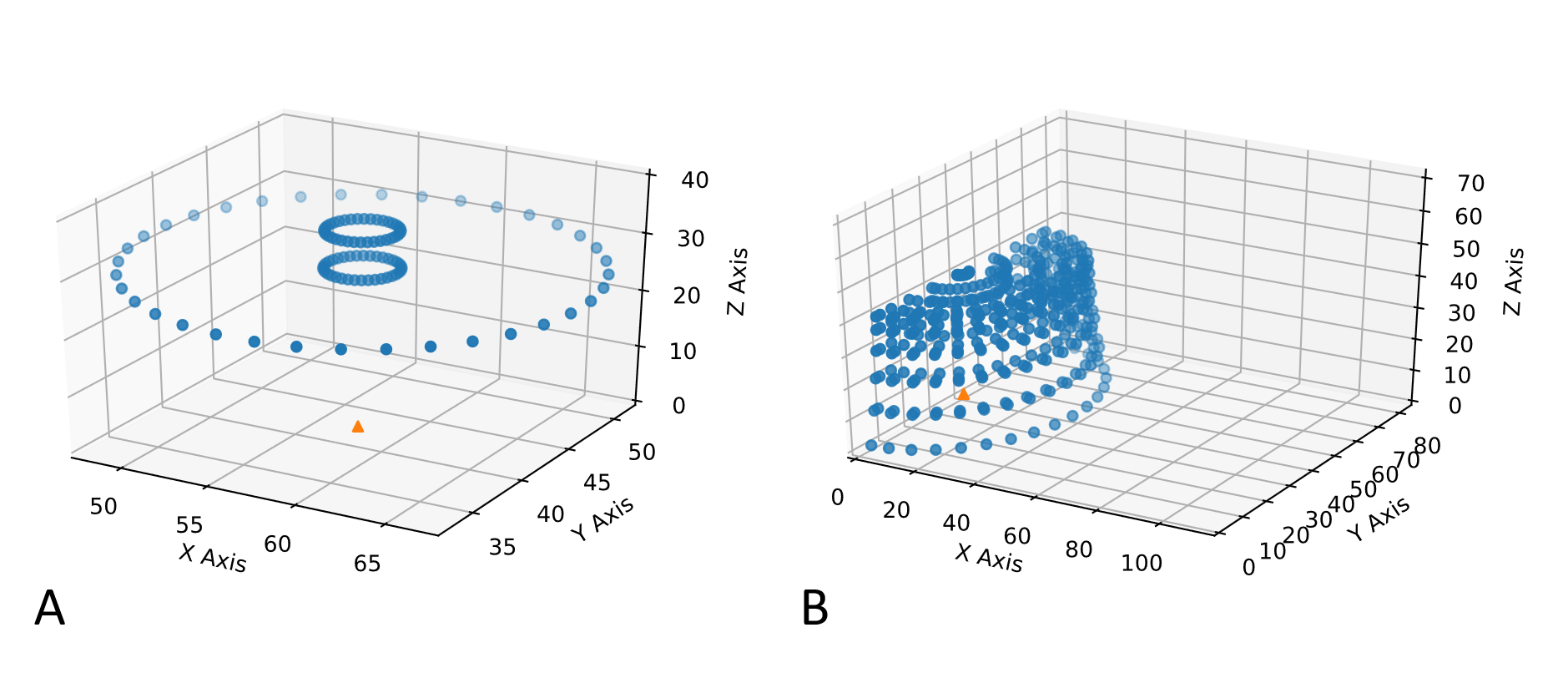}\end{center}

\caption{{\bf Camera positions}
The different camera positions from which the plant located
at the green triangle is imaged. Scenario A: Since the plant is located at the
border of the traversable volume, we have a cylindrical shape from
which we can image the plant. Scenario B: The plant is located in the
inside of the traversable volume, resulting in circles at different heights
and radii from which we image the plant.}
\label{fig6}
\end{figure}

Each image of the dataset is accompanied by two additional files.
The first is a copy of the original image that contains bounding boxes
corresponding to the cropped out subimages. The second is a JSON-file
containing the following metadata fields:
\begin{itemize}
\item \emph{version}:A version number differentiating file formats; this dataset's version is 1.5 and differs from 
earlier test sets in the number of data fields and formatting style.
\item \emph{file\_name}: A unique image identifier of the form \emph{yyyymmddhhmmss-pose\#.jpg},
where the first 14 digits encode year, month, day, hour, minute, and
second of when the image was captured. The number after \emph{pose}
denotes the position of a specific data-acquisition run.
\item \emph{bb\_file\_name}: A unique identifier for a copy of the master image with bounding boxes drawn on it. 
The format is equal to the one in \emph{file\_name}
but with a \emph{-bb} attached after the pose number.
\item \emph{date} and \emph{time}: Date and time at which the picture was
taken
\item \emph{room} and \emph{institute}: Abbreviated location of where EAGL-I 
was set up.
\item \emph{camera} and \emph{lens}: Information about the camera being
used. In the case that there is no specific lens information the
\emph{lens} field can be used for model information (in our case
we use \emph{camera} = GoPro and \emph{lens} = Hero 7 Black)
\item \emph{camera\_pose}: A literal containing the camera position in terms
of X, Y, and Z coordinates, polar-, and azimuthal angle.
\item \emph{bounding\_boxes}: A list of objects containing information for
all cropped subimages, containing the following fields for each such
image:
\begin{itemize}
\item \emph{plant\_id}: A unique identifier for each plant, consisting of
the first letters of its scientific name and a number, for example:
\emph{echcru002}
\item \emph{label}: The common name label, for example: \emph{BarnyardGrass}
\item \emph{scientific\_name}: For example \emph{Echinochloa crus-galli}
\item \emph{position\_id}: Denoting the positional ID at which the plant
was located
\item \emph{subimage\_file\_name}: A unique subimage identifier of the form \emph{yyyymmddhhmmss\#.jpg}, where \# is the position ID that ensures 
uniqueness
\item \emph{date\_planted}: The day we put the plant's seed in soil
\item \emph{x\_min, x\_max, y\_min, y\_max}: The subimage's coordinates
in the parent image, given as a percentage. A value of $x=0,\,y=0$
denotes the image's upper right corner, whereas $x=1,\,y=1$
denotes the lower left corner; this is conform to the
directions as defined in the OpenCV-library, which is used for our
image processing pipeline
\end{itemize}
\end{itemize}
Since the available imaging perspectives of a plant depends on where it is located,
we have sorted the position IDs into two classes: In the first class, four of the positions lie on
the edge of the volume that the gantry system can cover. That limits
the camera-poses from which we can image that position to half a cylinder. 
The second class of positions lie in the interior of the
coverable volume, resulting in a half-sphere of possible camera-poses
to image from. See Fig \ref{fig6} for a visualization
of the two different classes. The subimages are sorted by these two
location classes and saved into respective subfolders.

\subsection{Training a CNN}

Here we establish the value of data collected with the EAGL-I
system by training a CNN to sort plant images into one of two distinct classes.

\subsubsection{Model and Task}

The specific task we pose to the network is to differentiate between grasses
and non-grasses. As representatives for grasses we have barnyard grass,
wild oats and yellow foxtail. We chose this task (in contrast to other
classification challenges like identifying each species by itself
or for example differentiating the cash crop canola from weeds) for
two reasons: First, a significant portion of our training images includes
seedlings that have not grown their first true leaves, yet. Since
all non-grasses in our datasets are dicots, a visual distinction between
grasses and non-grasses is possible even during their earliest growing
stages.
Second, a key question to answer is how the data generation process
has to be improved such that models trained on the respective data
generalize to new scenarios. For this it is instrumental to test
the models on external data. Defining this rather general task allows
us to run the model with a wider variety of data, specifically to
plants that we did not have access to when generating the training
set. 

To perform this task we train a model based on the established ResNet
architecture \cite{he2016deep} with 50 layers and randomly initialized
weights. We average and normalize the input images to enhance the
actual differences between the pictures, which are the plants (in
contrast to the rather uniform blue background). To counteract the
slight imbalance between the two classes we introduce class weights
$c_{m}$ and $c_{d}$ defined as 
\begin{equation}
c_{m}=\frac{|\text{Total images}|}{|\text{Images of monocots}|},\,c_{d}=\frac{|\text{Total images}|}{|\text{Images of dicots}|}.
\end{equation}

\begin{figure}[!ht]

\begin{center}\includegraphics[width=0.8\columnwidth]{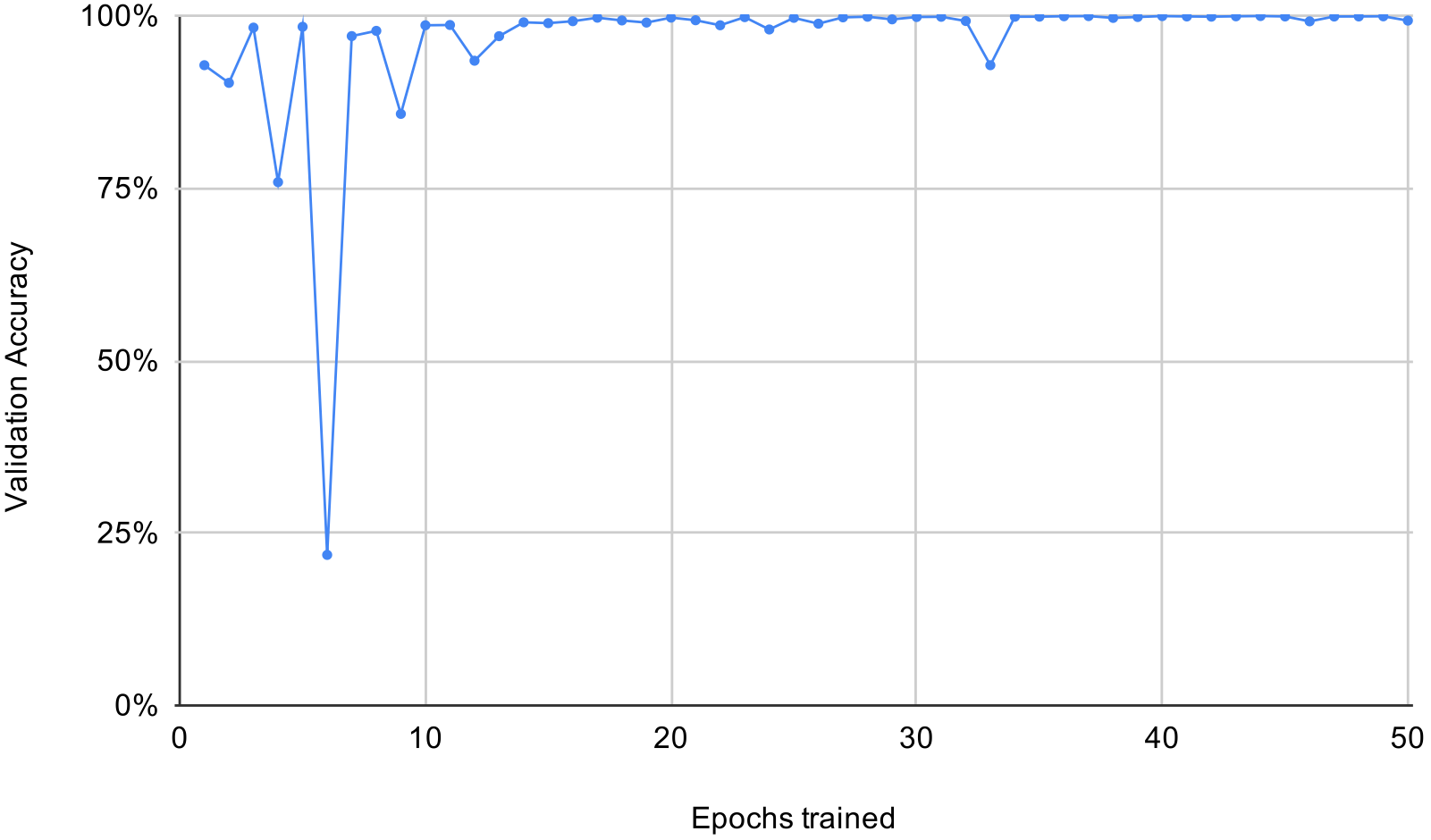}\end{center}

\caption{\bf{Validation accuracy in percent evaluated after each training epoch.}}
\label{fig7}
\end{figure}

We used 80\% of the data for training, reserved 20\% as validation data, and repeated
training over the entirety of the training set 50 times (each one
forming an \emph{epoch}). The validation accuracy achieved a satisfactory
convergence with a validation accuracy of 99.71\% after 50 epochs
(average of 99.79\% and a variance below 0.025\% over the last 20
epochs). The evolution of the validation accuracy per trained epoch is
graphed in Fig \ref{fig7}.

\subsubsection{Results in Different Scenarios}

To test our network's capabilities of distinguishing monocots from dicots, 
we presented it with new, previously unseen data. To investigate how
well the model generalizes, we tested 
it in the following scenarios that increasingly differ from the training data:
\begin{itemize}
\item Images of the same species taken by the EAGL-I system, but with new
individual plants. Those images differ from the training set only
in having different individuals of the same species. 
\item Images of the same species taken by the EAGL-I system, but under randomized
camera angles and distances.
\item Images of the same species outside EAGL-I's environment with a neutral
background taken by a smartphone camera.
\item A collection of Arabidopsis and tobacco plant images under lab conditions
produced by Minvervi et al. \cite{Minervini2015PRL}.
\item A collection of field data of sugar beets produced by Haug and Ostermann
\cite{haug15}.
\item A collection of plant seedling images produced by Giselsson et al.
\cite{Giselsson2017}.
\end{itemize}
The results for the different scenarios are given in Table \ref{tab:different-scenarios}.

\begin{table}[!ht]
\caption{{\bf Test datasets}}
\begin{center}
\begin{tabular}{|>{\centering}p{3.5cm}|c|c|c|}
\hline 
Scenario & Size of Test Set & Correctly Identified & Accuracy\tabularnewline
\hline 
\hline 
EAGL-I camera, same species, same angles & 3494 & 3437 & 98.37\%\tabularnewline
\hline 
EAGL-I camera, same species, randomized angles & 520 & 513 & 98.65\%\tabularnewline
\hline 
Neutral Background, smartphone, same species & 56 & 50 & 89.29\%\tabularnewline
\hline 
Minervi et al. \cite{Minervini2015PRL} & 347 & 283 & 81.56\%\tabularnewline
\hline 
Haug and Ostermann \cite{haug15}, field data, sugar beets & 162 (of 494) & 120 & 74.07\%\tabularnewline
\hline 
Giselsson et al. \cite{Giselsson2017} field data, different species & 500 (of 5539) & 316 & 63.20\%\tabularnewline
\hline 
\end{tabular}
\end{center}
\label{tab:different-scenarios}
\end{table}

In the first scenario an accuracy of 8.37\% was achieved, indicating
that the model generalizes to new plants of the same species imaged
under the same circumstances. The model we used has converged on the
training data and might even show first signs of 
overfitting. 
 For example, if we apply the model that is available after 40 
epochs of training, the accuracy, the accuracy
on the test data increases by 0.5\% to 98.89\%. This is a sign that
improving the accuracy further requires a richer dataset, a more
complex model, or a combination of both. 

When we randomize the positions from which we take images, we see
that it has no significant impact on the model's overall accuracy.
From this we conclude that the variety of angles covered in our
training sets are sufficient for the model to be insensitive to
imaging angles (such as profile shots or overhead shots) when
distinguishing grasses from non-grasses. 

For images taken by smartphone with a neutral background, 
a high accuracy above 89\% is still achieved. The model generalizes to new imaging conditions, then,
although with reduced accuracy (which is to be expected).
However, test data is significantly smaller in quantity, since
its generation required to manual capture and labeling. 
Thus, the accuracy on the test data could deviate from the model's accuracy on a larger set of similar images. 
To give a more complete picture of where the model's true accuracy lies, we calculated 
a Clopper-Pearson confidence interval of $[0.781,0.959]$ at a confidence level $\alpha=0.05$.

We now explore how our model generalizes to data produced by others for 
species that are not represented in our training set. 
The dataset in
\cite{Minervini2015PRL} consists of 283 images of Arabidopsis plants
and 62 tobacco plant images. The images are all taken top-down
and show the plants at different growing stages. The dataset was taken
with phenotyping applications in mind and contains images of
dicots only.
On the overall data we achieve an accuracy of 81.56\%, which in this
case coincides with how many plants were classified as dicots. This is a strong demonstration that 
our model can  generalize to species not included in the training data. If we break the test data down via
the two species, we see that the model has an even better performance
on the Arabidopsis images (91.23\%), while performing rather poorly
on tobaccos (37.1\%). This tells us that the training set we generated is missing
dicots that are morphologically similar to tobacco plants, and that
we need to include these to achieve a more robust model. 

As a next step to test how far our binary classifier generalizes, we
applied it to the dataset provided in \cite{haug15}. This dataset
consists of field data taken in a sugar beet field and features crop
and weed plants. Since the annotations do not specify the weeds, we only use images that
show sugar beets (a dicot). The original
data in \cite{haug15} shows several plants per image. Thus, we used the metadata
provided by the authors to crop out the sugar beet plants. Still, on
many of those cropped images we see plants overlapping into the cropped section. 
This is a challenge for our model, which was trained on single
plant images only. The test data also features natural background (dirt) in
contrast to the rather homogenous backgrounds on images we trained
and tested on before. On the aforementioned subset our model achieves an
accuracy of 74.04\%. While not perfect, this shows that the model
has already some capacity to generalize to new lighting and background
conditions and another species of plants the model has not trained on.

\begin{figure}[!ht]

\begin{center}\includegraphics[width=1.0\columnwidth]{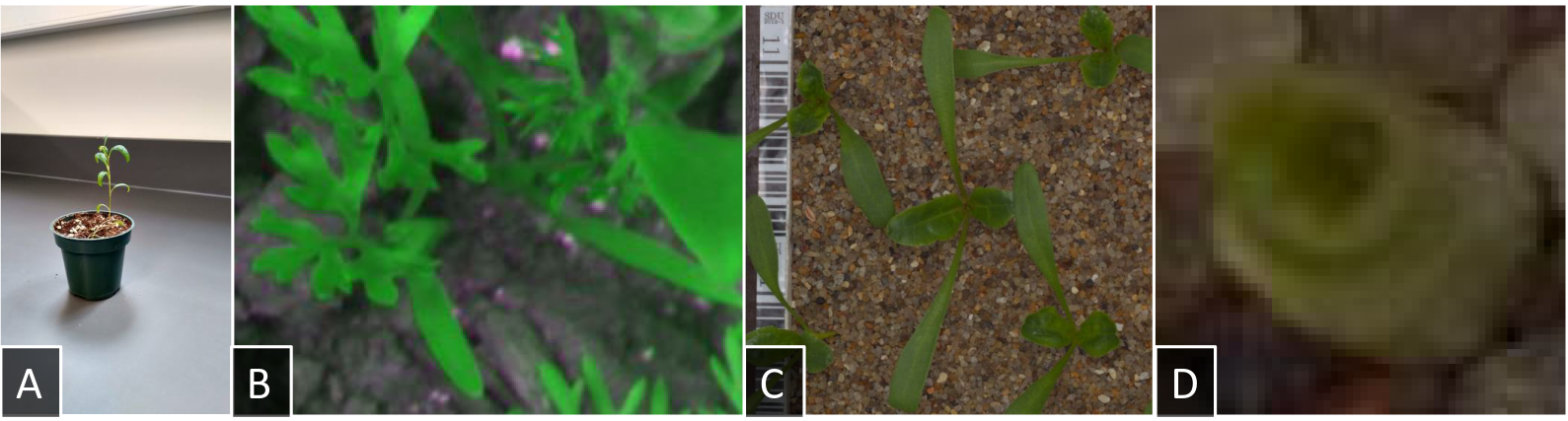}\end{center}

\caption{{\bf Examples of test data} 
A: A test image taken with a smartphone. B: A cropped out image from the dataset in \cite{haug15}.
C-D: Two examples from the seedling dataset of \cite{Giselsson2017}: a high-res image containing multiple sugar beets and artifacts to the left border and a low-res image
containing a maize seedling.}
\label{fig8}
\end{figure}

Finally, we applied our model to the dataset given in \cite{Giselsson2017}.
This dataset is very challenging for various reasons: First, the contrast
between plant and background is not as distinguished as in our training
set or the other test sets. Second, the data contains many plants at their earliest
growing stages and as a result some images have a resolution
as small as 49 x 49 pixels (see Fig \ref{fig8}
for an example of a high- and low-resolution image). Third,
as in the previous test dataset, the images contain multiple plants, though the authors of \cite{Giselsson2017} have ensured that only one species
is present in each image. Fourth, the dataset contains only species
that are not present in our training data.
Still, our goal to distinguish monocots from
dicots remains. To this end, we sorted the plants in \cite{Giselsson2017} into
two categories: maize, wheat, blackgrass, and loose silky-bent represent 
monocots; whereas sugar beet, mayweed, chickweed, shepherd's purse, cleavers, charlock, fat hen, and cranesbill
comprise the set of dicots. To test the model we chose the 250 images
with highest resolutions for both classes. The achieved accuracy is
63.2\% (confidence interval $[0.588,0.674]$ at $\alpha=0.05$). Although
this value does not lie far above 50\%, it is still significant as
it shows that the model generalizes to some extent to data that shares only small similarities to the training data. 
A first step to improve accuracy would
be to detect and crop out the plants in the test data before
classification. This reduces the number of artifacts
and ensures that no multiple plants are in a single image. Another improvement
for this specific test data would be achieved by generating training
data more suitable to the task, meaning imaging species
used in Ref. \cite{Giselsson2017} and focusing on overhead shots. As presented
in Subsection \ref{subsec:postprocessing}, the blue background in
the training images can be replaced by images of the granulate appearing in the images of Ref. \cite{Giselsson2017} to achieve an even higher similarity to the test data. 
This idea to create training data that resembles the data we can expect in an application is exactly
the raison d'etre of the EAGL-I system.

\section{Conclusion and Future Work\label{sec:Conclusion-and-Future}}

In this paper we described the construction, operation, and utility of an embedded system (EAGL-I) that can automatically generate and label large datasets of plant images for machine learning applications in agriculture.  
Human interaction is reduced to selecting the plants to image and placing them inside the system's  imaging volume. 
EAGL-I can create a wide diversity of datasets as there are no limitations in plant placement, camera angle,
or distance between camera and plant within this volume. 
Furthermore, the use of blue keying fabric as a background enables additional image processing techniques such as background removal and image segmentation.
The system's performance was
demonstrated along several dimensions.  With a subimage production time of $\sim 4.8$~s, we produced a dataset of over 34,000 labeled images of assorted weeds that are common in the Province of Manitoba.  We subsequently used that dataset to
train a simple convolutional neural network for distinguishing monocots from dicots, which in turn was tested on a variety of other datasets with quite favorable results. 

We see the EAGL-I system as a important stepping stone to enabling new ML-based technologies in agriculture, such as automated weeding,  that will  require large amounts of labeled training data.
Our system also provides opportunities to follow research
questions that were not accessible before. For example, 
with the ability to generate a quasi-unlimited source of data ourselves, we can investigate
how quantity and quality of training data influences machine learning models. 
Normally the amount of training data for a problem is
hard-capped and acts as an observation limit for this type of research. 

There are many other directions for improvements and future work for the EAGL-I system, of which
we mention a few here. 

Lighting: The addition of programmable LED lighting elements are being
planned and  will allow us to customize lighting
conditions on a per image basis, if desired.  This will enable an even wider variety of images to be collected
by simulating different lighting scenarios, e.g.\ sunny, cloudy, evening
hours, etc. 

System design and dimensions: 
EAGL-I is presently limited to take images inside its coverable volume putting hard limits 
on the number and size of plants that can be imaged in a given run. This leads to research questions about the 
design of imaging systems that are specific for the creation of labeled data. The challenge, then,  is to design a system that can produce
a wide variety of images---preferably including a wide variety of plants differing in size and growing pattern---at a small cost and high imaging rate.  The gantry architecture of EAGL-I is simple and functional, but may not be optimal. One direction we are considering is mounting  linear actuators and cameras directly to the walls and ceiling of a growth chamber.

Three dimensional plant data: Since we have full control over the
camera position, we should be able to use software to reconstruct 3-dimensional plant
models from 2-dimensional images taken from different angles. This could be
a simple depth map extracted from two or more images via parallax or
a 3d-point cloud combining more images. Alternatively, we can mount
different imaging systems, such as stereoscopic cameras, to the gantry head in order to generate 3d data directly.

Detection and imaging of plant organs:  Often one is interested in the
specific parts or organs of a plant, such as wheat spikes.  To image these effectively, we have to solve how to point 
the camera at the desired organ for each plant. To achieve this we could combine machine
learning techniques with our imaging system to bootstrap a training
dataset for identifying specific plant organs. From there we can use a model
to automatically move the camera in close proximity of the wheat spikes, say, and capture high resolution
images. Both, the training set for identification, and the image dataset of high resolution wheat spikes would be valuable
for subsequent applications such as phenotyping, blight detection and crop evaluation in the field. 

\section{Data Availability}
The dataset and model described in Section \ref{sec:The-Weed-Dataset} are publicly available \cite{weedling2020}.
The production of much larger future datasets is underway and will include Canadian crop plants, such as wheat, canola, soybean, and pulses.  
We presently envision depositing these datasets at the 
Federated Research Data Repository\footnote{\url{https://www.frdr-dfdr.ca/repo/}} through a data management plan developed with the tools provided by the Portage Network \footnote{\url{https://portagenetwork.ca}}.

\section{Acknowledgments}
The authors would like to the thank the following people for their support, generosity and vision:  Ezzat Ibrahim for establishing
the \textit{Dr.\ Ezzat A.\ Ibrahim GPU Educational Lab} at the University of Winnipeg, which we used extensively for the computing resources needed here; Rafael Otfinowski, Karina Kachur and Tabitha Wood for providing us with seeds, plants and laboratory space to develop our prototypes and datasets; Jonathan Ziprick for many helpful conversations about the gantry system and actuators; and Russ Mammei and Jeff Martin for allowing us to use their magnetic field mapping system as the first prototype of EAGL-I.


\end{document}